# Automatic detection of glaucoma via fundus imaging and artificial intelligence: A review

Lauren Coan (MSc, School of Computer Science and Mathematics, Liverpool John Moores University, UK)

Dr. Bryan Williams (PhD, Lecturer in Biometrics and Human Identification, School of Computing and Communications, Lancaster University, UK)

Mr. Krishna Adithya Venkatesh (Research associate, Department of Glaucoma, Aravind Eye Hospital, Pondicherry, India)

Dr. Swati Upadhyaya (DNB, Medical Consultant, Department of Glaucoma, Aravind Eye Hospital, Pondicherry, India)

Dr. Silvester Czanner (PhD, Subject Head Computer Science and Mathematics, School of Computer Science and Mathematics, Liverpool John Moores University, UK; and Researcher at Faculty of Informatics and Information Technologies, Slovak University of Technology, Slovakia)

Dr. Rengaraj Venkatesh (DNB, Medical Consultant, Department of Glaucoma and Chief Medical Officer, Aravind Eye Hospital, Pondicherry, India)

Professor Colin E. Willoughby (Professor of Ophthalmology, Biomedical Sciences Research Institute, Ulster University, UK)

Dr. Srinivasan Kavitha (MS, Head of the Department, Department of Glaucoma, Aravind Eye Hospital, Pondicherry, India)

Dr. Gabriela Czanner (PhD, Reader in Statistics for Healthcare, School of Computer Science and Mathematics, Liverpool John Moores University, UK; and Researcher at Faculty of Informatics and Information Technologies, Slovak University of Technology, Slovakia).

**Abstract**

Glaucoma is a leading cause of irreversible vision impairment globally and cases are continuously rising worldwide. Early detection is crucial, allowing timely intervention which can prevent further visual field loss. To detect glaucoma, examination of the optic nerve head via fundus imaging can be performed, at the centre of which is the assessment of the optic cup and disc boundaries. Fundus imaging is non-invasive and low-cost; however, the image examination relies on subjective, time-consuming, and costly expert assessments.

A timely question to ask is can artificial intelligence mimic glaucoma assessments made by experts. Namely, can artificial intelligence automatically find the boundaries of the optic cup and disc (providing a so-called segmented fundus image) and then use the segmented image to identify glaucoma with high accuracy.

We conducted a comprehensive review on artificial intelligence-enabled glaucoma detection frameworks that produce and use segmented fundus images. We found 28 papers and identified two main approaches: 1) logical rule-based frameworks, based on a set of simplistic decision rules; and 2) machine learning/statistical modelling based frameworks. We summarise the state-of-art of the two approaches and highlight the key hurdles to overcome for artificial intelligence-enabled glaucoma detection frameworks to be translated into clinical practice.

# 1. Introduction

Glaucoma is one of the leading causes of global vision impairment [A] and the second most common cause of blindness globally (72). By 2040, it is estimated that 112 million individuals globally will suffer from the disease (76). With the ageing global population rising (72), there will be a corresponding increase in glaucoma cases that will continuously challenge our resources worldwide (62). The global burden of vision impairment and/or blindness due to glaucoma is significantly associated with a decrease in quality of life, physical functioning, and mental health (15). Although irreversible, early diagnosis of glaucomatous neuropathy allows for treatment to be implemented which may slow or prevent glaucoma progression and blindness.

Currently, in the United Kingdom (UK), glaucoma detection is opportunistic, most frequently accomplished by optometrist assessment in the community (34). Around half of the glaucoma patients in the community remain undiagnosed (12). A recent population-based study in Northern Ireland suggests that the majority of people with glaucoma are undetected and two-thirds of glaucoma patients within the study were unaware of having the disease (51).

Although a worldwide problem, the burden of glaucoma is higher within developing countries (22) and the disease disproportionately affects African and Asian countries (64). Moreover, studies indicate that more than 11.2 million individuals in India are affected by glaucoma, constituting approximately one-fifth of the global burden of the disease (68). In the UK, hospital eye services (HES) are the busiest outpatient service in the National Health System (NHS) and are responsible for 8.3% of all outpatient activity[B]. Glaucoma accounts for 25% of HES appointments. Individuals with, or at risk of, glaucoma are detected by community optometrists and referred to HES, 15-20% of the new referrals will have glaucoma and around 50% will be discharged at the first visit, costing the NHS upwards of £75m/year (7).

Given this worldwide problem of glaucoma detection, the urgent question is how close we are to having accurate artificial intelligence (AI) enabled glaucoma detection (34) and whether such AI can then be explained to the clinician and patient. The answer to this question is two-fold: we need to understand the process of detecting glaucoma in clinical practice, and then we need to determine if artificial intelligence can accurately detect glaucoma while also providing key explanations, mimicking the clinician's reasoning.

**The detection of glaucoma by a clinician**

The presence of glaucoma instigates changes in the structure of the optic nerve head (ONH) (Figure 1a) and retinal nerve fibre layer (RNFL) associated with functional visual defects. Structural changes are manifested by a slow, yet progressive, narrowing of the neuroretinal rim, indicating degeneration of retinal ganglion cells axons, and astrocytes of the optic nerve (9). To evaluate the narrowing of the neuroretinal rim (NRR) the clinician needs to identify the boundary contours of the cup and disc. Such contours then help when explaining to the patient the reasoning behind the diagnosis, and thus help the patient to participate in the discussion and treatment decision. Given the significance of patient involvement in the decisions regarding their care and the importance of AI explainability, this review focuses on AI that provides optic cup (OC) and optic disc (OD) contours.

[Figure 1 (a,b) here]

*Figure 1: Fundus photograph examples (a- left) with labels of the optic nerve head and (b) with (Inferior-Superior-Nasal-Temporal) ISNT quadrants.*

Glaucoma detection is a challenging and lengthy process, relying on multiple examinations and clinical expertise. The National Institute for Health and Care Excellence (NICE) in the UK recommends examination of the ONH via a technique called fundus imaging (15). In addition, many clinicians use Optical Coherence Tomography (OCT), which can provide three-dimensional information to aid diagnosis, but this is difficult to interpret, requiring clinical expertise.

ONH assessment via imaging modalities evaluates structural changes in the ONH. Such structural changes often precede the development of visual field loss (73). Fundus cameras are advantageous due to their relatively low cost compared to their imaging counterparts such as OCT and Heidelberg Retinal Tomography (HRT). Yet, they provide images that are of suitable quality to detect abnormalities in the ONH for evaluating ocular health (78). Due to their cheaper cost, fundus cameras are readily available in a range of settings from rural community centres, local ophthalmologist stores, and hospitals. Although in recent years, OCT imaging has become cheaper and more widely available to optometrists in economically stronger countries, low-cost portable fundus cameras have been developed that can be more readily utilised for wider population-based screening of glaucoma in lower resource settings or isolated communities. Therefore, this review focuses on AI that utilises fundus images.

**Detecting glaucoma via AI**

AI is a computer system that can perform tasks that normally require human intelligence such as the task of glaucoma detection via examination of a fundus image. AI works by combining data and expert knowledge into fast and intelligent computer algorithms (64). AI is an umbrella term that encapsulates machine learning algorithms, which in turn include deep learning (DL) methods (53). In recent years, we have seen a significant increase in the utilisation and development of AI, alongside momentous developments in technology. Automated algorithms are already being used in some clinics including ophthalmology (4) such as the FDA approved AI-based device that detects diabetic retinopathy.

Technological advances mean that the creation of AI-enabled glaucoma detection methods via the modality of fundus images is a realistic proposition (55). Several portable fundus cameras have been developed, such devices are small, inexpensive and are becoming straightforward enough to be administered by laypersons (38). A recent review on the use of telemedicine in glaucoma highlights that machines that are less operator dependent should give more objective results even when they are operated by less experienced personnel at remote sites (45).

If AI-enabled glaucoma detection methods using fundus imaging could be deployed in screening mechanisms, this could aid with reducing human error (e.g., observer bias and fatigue) and be used for large-scale screening at a low cost. This could provide much-needed eye care services to remote rural areas, particularly in nations where there is a scarcity of qualified, skilled, and competent ophthalmologists (50). In the near future, automated image interpretation for screening, referral decision-making, and patient monitoring is likely to play a crucial role in frontline eye care. Even in resource-rich care settings such as the NHS in the UK, referral refinement with AI has the potential to address the staggering outpatient appointment demand while reducing false positive referral rates[c] .

What remains unclear is the full state of AI-enabled glaucoma detection, namely the frameworks that utilise fundus cameras while providing the contours of the OC and OD. To understand the potential application of AI-enabled glaucoma detection we must first answer many questions (i.e., how accurate are the AI methods, how suitable/appropriate are they, how have they been trained/tested/validated). Following this, we can then identify the next steps to further develop AI-enabled glaucoma detection

**Overview**

The key objectives for the review are: (1) To outline and clarify the main AI terminology used with AI-enabled glaucoma detection such that the review is accessible to ophthalmologists, (2) To provide a detailed overview of the state-of-art AI-enabled glaucoma detection methods that use segmented fundus images - highlighting the two approaches used when using fundus imaging, and (3) To provide a discussion on the progress of AI-enabled glaucoma detection methods and highlight areas that require further work.

In the following sections, we provide a clinical and technical background and define the terminology referred to throughout this review. Section 3 then defines the methods used for the literature search and outlines the key information extracted from the reviewed papers. Section 4 explains the methods employed in this review and Section 5 covers the results of the review. Lastly, Section 6 provides a discussion, conclusions and future work recommendations.

## 2. Clinical terminology and brief background

### 2.1. Cup to disc ratio

The cup-to-disc ratio (CDR) is a universally acknowledged parameter for describing glaucomatous neuropathy, obtained from assessment of the ONH. There are different variants of the CDR parameter however, the primary two are the vertical cup-to-disc ratio (vCDR) and the area cup-to-disc ratio ACDR.

The vCDR is defined as:

$$vCDR = \frac{Vertical\ Cup\ Diameter}{Vertical\ Disc\ Diameter}$$

The ACDR is defined as:

$$ACDR = \frac{Area\ of\ Cup}{Area\ of\ Disc}$$

Although well used in practice, the CDR parameter is limited in cases of genetically large or small OD, large OC cases, and in cases where myopic ONH changes are present (52,14); in such instances, the CDR can be misleading (33) and lead to errors in diagnosis. Other morphometric features such as the rim-to-disc ratio (RDR) and horizontal cup-to-disc ratio (hCDR) can also be considered. In contrast to the CDR, a decrease in the RDR indicates glaucomatous neuropathy. The ACDR provides a 2-D feature-based measurement allowing structural changes of the ONH to be assessed.

### 2.2. Neuroretinal rim area ratio

The Neuroretinal Rim (NRR) is the area between the OC margin and the OD margin which comprises retinal nerve fibre axons. When using fundus images, the NRR is the area left behind when subtracting the OC from the OD. The NRR is divided into four quadrants: inferior, superior, nasal, and temporal as shown in Figure 1b.

The NRR area (75) is calculated as:

$$NRR = \frac{Area\ in\ Inferior\ Quadrant + Area\ in\ Superior\ Quadrant}{Area\ in\ Nasal\ Quadrant + Area\ in\ Temporal\ Quadrant}$$

The four quadrants of the NRR are typically expected to satisfy the Inferior-Superior-Nasal-Temporal (ISNT) rule (I>S>N>T) (52). Whilst the cup-to-disc ratio parameter focuses on the OC size with respect to the OD, the ISNT rule focuses on the NRR width i.e., the area between the boundary of the OC and OD (52). The ISNT rule follows that the inferior rim is thicker than the superior rim, which is thicker than the nasal rim, which is thicker than the temporal rim in a healthy eye (17). Any violation of the

ISNT rule can be seen as a sign of glaucomatous neuropathy. However, this is not always the case (i.e., a healthy NRR can violate the rule) (75). As such, the ISNT rule is not recognised as a diagnostic test but rather a clinical tool.

## 2.3. Disc Damage Likelihood Scale

The Disc Damage Likelihood Scale (DDLS) is a grading protocol that divides glaucomatous progression into ten stages while accounting for OD size (14,71). This method has proved to be time-consuming, requiring a detailed grading protocol with a standard set of images for comparison purposes. Also, it necessitates further training of clinicians. The advantage of this method is in higher inter-observer repeatability (73) and higher agreement with the gold standard (14) than the vertical CDR.

# 3. Technical terminology and brief background

Within the AI community, many terms are used interchangeably; we define the key terminology used throughout this review.

## 3.1. Fundus image segmentation

In medical image processing, image segmentation refers to the (typically automated) partitioning of an image into multiple clinically meaningful segments. Fundus image segmentation is the process of finding the visible boundaries (or "contours") of the OC and OD. Manual image segmentation can involve a trained expert, such as a clinician or grader, manually annotating the boundary of the OC and OD. Whereas automatic image segmentation is accomplished by mathematical algorithms. To date, there has been a large number of AI methods proposed for automatic image segmentation of the ONH. Popular approaches include level-set-based algorithms, threshold-based algorithms, and clustering-based algorithms (6,79). The resulting annotation of the boundaries is what we call the **segmented image**.

[Figure 2 here]

*Figure 2 - Examples of the automatic optic cup and disc segmentation in fundus images centred on the optic nerve head. The yellow line represents the optic cup boundary, and the blue line represents the optic disc boundary.*

## 3.2. Image features

In the AI community, the term "image feature" refers to a variable or parameter derived from an image. Two types of image features can be extracted from fundus images; namely clinically interpretable features and abstract features.

**Clinically interpretable features** are features with clinical meaning (e.g., vCDR and NRR area). These clinical features have been developed over many years by expert ophthalmologists and can be intuitively explained to a patient. In contrast, we can also consider mathematically derived **abstract features**. Such features may not be clinically interpretable due to being constructed via a mathematical or statistical process.

## 3.3. Probability of glaucoma

In general, AI calculates a probability of glaucoma for an unseen new fundus image as a number between 0 and 100% (e.g., 90%). This probability is interpreted as follows: given the training set that AI used and the mathematical/statistical method that the AI is built on, the AI believes that the chance of glaucoma is 90%, i.e., among the 10 images that look like the new image, 9 do have glaucoma and 1 do not. The value of the probability of glaucoma should be calculated to reflect the prevalence in the population of interest via e.g., Bayesian updating rule. If the probability provided by AI is 50%, then the AI is not certain if the new image is glaucomatous or not. However, if the probability is 99%, this does not mean that AI is certain that it is glaucoma. For example, if the new image is not represented well in the training dataset, then AI is not sufficiently trained to judge the new image, and therefore it should be able to express its uncertainty[D]. The calculation of uncertainty of AI is a complex problem and it is a current area of intensive research.

## 3.4. Image classification

We use the term "image classification" to refer to the automated process of determining the category to which a given fundus image belongs e.g., healthy or glaucomatous group (binary classification); or healthy, suspected glaucoma or glaucoma group (multi-class classification). This process is also referred to as image discrimination (16) or disease prediction. To achieve the classification, AI can apply a threshold to the estimated probability of glaucoma, e.g., if the image's estimated probability is higher than the threshold, the image is classified as glaucoma. If AI is uncertain in the calculated probability then such uncertainty will propagate into the uncertainty of the classification.

## 3.5. Classifier

We use the term "classifier" to refer to a mathematical or statistical or machine learning method used within the AI framework to estimate which disease category the patient belongs to (glaucomatous, suspected glaucomatous or healthy). Popular classifiers are support vector machines or logistic regression.

## 3.6. AI framework

We use the term "AI framework" to encapsulate the whole process of automatically classifying a given fundus image into a group (glaucomatous, suspected glaucomatous or healthy). This process can comprise many steps including (but not limited to) image segmentation, feature extraction, and using the image features (via various methods) for discrimination of glaucomatous neuropathy. The framework's final step is to provide the classification output for a given image.

**One-step AI framework.** Some AI frameworks do not require and do not produce segmented images. They learn a link between the fundus images and the disease status and then directly provide their estimate of the disease group. To build such AI, a so-called **end-to-end image classification** method is needed, which is computationally intensive as images have millions of pixels. Such computation can be enabled via DL algorithms (47) (e.g., convolution neural networks). This is possible due to their complex interior working architectures with complex transformations across multiple layers.

The complexity of DL makes such frameworks hard to explain. For this reason, they are commonly referred to as 'black box methods', meaning they inherently do not provide an understanding of how they learn i.e., they do not explain their inner working (also known as "not explainable"). By their nature, DL methods do not provide intuition as to why they estimated glaucoma for a particular fundus image i.e., they do not explain the output for a clinician or patient (77) (also known as "not interpretable"). Whilst this can be acceptable in low-stake applications, interpretable and explainable AI is crucial in the medical domain. The AI methods proposed must foster trust from clinicians, regulators and patients alike. To do this, researchers should be able to present why their AI has provided the respective output [DE]. It should be noted, this underlying requirement of interpretable and explainable AI does not have to come at the cost of accurate AI (65).

**Two-step AI framework.** Other AI frameworks produce a segmented image as the first step. In this step, the segmented image can provide clinically interpretable features (e.g., CDR ratio and NRR area), or abstract features (e.g., texture and colour features). The second step then uses such features and provides an estimate of the disease group. In general, these two-step frameworks have increased interpretability as they provide the clinician and patient with the segmented image which allows demonstration of the part of the image leading to the AI's output for a patient (40) and facilitates further investigation. Such approaches were first used in ophthalmology by De Fauw in September 2018 (19) and MacCormick in January 2019 (49).

### 3.7. Evaluating the performance of AI

Careful evaluation of AI is required to understand the AI's performance capabilities; that is, how well the AI agrees with the gold standard. By the "gold standard" (also referred to as ground truth), we refer to the decision of a clinical expert on whether the eye has glaucoma or not. There is no single measure that alone would be enough to evaluate the performance of AI. Hence, a combination of measures is required to give a complete overview of the AI framework's capabilities. In what follows we briefly mention the most important measures for evaluating the performance of AI.

**Confusion matrix**. The confusion matrix (Table 1) is used to give an overall representation of the performance of the AI's framework. Using this confusion matrix, key performance metrics are derived.

[Table 1 here]

The true positives ($T_P$) are the glaucomatous observations that have been correctly classified, whereas the true negatives ($T_N$) are the non-glaucomatous observations that are correctly classified as non-glaucomatous. The false positives ($F_P$) are the non-glaucomatous observations that are incorrectly classified as glaucomatous, and the false negatives ($F_N$) are the glaucomatous observations that are incorrectly classified as non-glaucomatous.

The accuracy metric is the proportion of correctly classified images. Sensitivity (aka true positive rate) is the proportion of actual positive cases (i.e., glaucomatous) that are classified as positive. Specificity (aka recall) is the proportion of actual negative cases (i.e., healthy) which are classified as negative.

The positive predictive value (PPV) is the probability that an individual with a positive reference test truly has the disease whilst the negative predictive value (NPV) is the probability that an individual with a negative reference test truly does not have the disease.

$$Sensitivity = \frac{T_P}{T_P + F_N}$$

$$Specificity = \frac{T_N}{T_N + F_P}$$

$$Accuracy = \frac{T_P + T_N}{T_P + T_N + F_P + F_N}$$

$$Positive\ Predictive\ Value\ (PPV) = \frac{T_P}{(T_P + F_P)} \times 100$$

$$Negative\ Predictive\ Value\ (NPV) = \frac{T_N}{T_N + F_N} \times 100$$

False positives are mistakes that potentially could lead to unnecessary further testing/referrals. Arguably false negatives are more serious in glaucoma as the disease is not identified and treated at the earliest stage. The detection of glaucoma would then occur at later stages, resulting in advanced and irreversible ONH damage and possible visual field loss, impacting the patient significantly. To this end, an effective framework (with high sensitivity) for the detection of potential glaucomatous subjects at the earliest stage is paramount.

**Area under receiver operating characteristic curve (AUROC).** A Receiver Operating Characteristic (ROC) curve plots the true positive rate ($sensitivity$) vs the false positive rate ($1 - specificity$) at all classification thresholds. The AUROC is defined as the area under the ROC curve. If we are presented with a pair of eyes, one with glaucoma and one without glaucoma, then the AUROC metric is interpreted as the probability of correctly distinguishing the glaucomatous eye from the non-glaucomatous eye. An AUROC of 0.5 is the equivalent to the flip of a coin.

**Internal and external evaluation of AI.** AI methods are tested to compute the aforementioned performance metrics (i.e., accuracy, sensitivity, AUROC etc). AI must be evaluated on data that has not been used within its training component. There are two methods for evaluating AI: internally and externally. In internal evaluation, the dataset can be split into two partitions, one is used for training and one for testing (e.g., 70:30 split). Hence, an image can be either be in the training or testing set, but not in both.

Another approach to internal evaluation is k-fold cross-validation. When using k-fold cross-validation, the dataset is randomly split into $k$ equally sized partitions; $(k - 1)$ partitions are used for training the classifier and the final partition is used for testing. This is repeated $k$ times with the performance metrics being retained each time. The final metric presented is the average of the $k$ splits. Generally, the value of k is set to five or ten for optimal bias-variance trade-off. Such evaluation approaches are called internal as all images come from the same source (i.e. the same cohort), and hence it may not be sufficient for evaluating the generalisability of the AI.

Conversely, external evaluation consists of testing the framework on data from a different source (then the data used for training). This could be a dataset acquired from a different cohort or device. Whilst internal testing gives insight into the performance capabilities of the framework, external testing is required as it provides an understanding of the generalisability of the framework with unseen data from different sources.

## 3.8. Reporting guidelines for AI in healthcare

With the ongoing developments of AI for health applications, there has been an increase in published guidelines for the reporting of the methods. The key information that should be reported includes the imaging device, contextual study setting, detailed cohort information and data processing methods (25). With the use of AI, further detail is required to be reported comprising the technical aspects of the methods presented. Recently, new standards specific to reporting studies of machine learning/AI interventions have been in development. This includes TRIPOD-ML, SPIRIT-AI and CONSORT AI (25) under the EQUATOR initiative [F].

## 4. Methods

We performed a comprehensive literature search, details of which can be found in the Method of Literature Search section. A table was used to extract all relevant information from the selected papers. For this review, we extracted information regarding the author, year of publication, approach to classification, data used (sample size, availability of the data publicly, number of data annotators, imaging device details), techniques used for segmentation, validation techniques applied, performance metrics of the methods (accuracy, sensitivity, specificity and AUROC). Whilst this review is primarily focused on assessing the classification of glaucoma following segmentation, we do provide details about methods to segmentation as this is a key step in the pipeline and can heavily influence classification results.

# 5. Results

## 5.1 Papers included

We identified a total of 1018 papers (Figure 3) to meet keyword search (Section 7). After the removal of 252 duplicates, papers were screened based on titles and abstracts. 599 papers were removed due to unsuitability for this review. Then, papers were screened based on text. From this, 139 papers were removed due to unsuitability for this review. There were three main reasons papers were labelled as unsuitable in this review (from most prevalent): (1) they proposed a one-step AI framework that did not require any segmentation of the fundus images, (2) they focused purely on segmentation and provided no framework for classification of glaucomatous optic neuropathy, (3) they did not utilise the modality of fundus images or they combined fundus images with other imaging modalities. The final number of papers found to meet eligibility criteria (Section 7) for this review was 28.

[Figure 3 here]

*Figure 3: Flow diagram of papers included within the review.*

## 5.2 Characteristics of identified papers

We have highlighted two distinct approaches to the classification of glaucoma from segmented images. We termed the first approach the logical rule-based framework due to the use of straightforward threshold rules (IF-ELSE statements) based on clinically interpretable imaging features. The second is machine learning/statistical modelling frameworks which exploit the imaging features in a range of classification models/algorithms for glaucoma detection. In this review, 11 papers were identified as using the logical rule-based framework, whilst 17 papers used machine learning/statistical modelling frameworks.

[Figure 4 here]

*Figure 4: Pathways of frameworks for two-step AI-enabled glaucoma detection*

## 5.3 Logical rule-based AI frameworks for glaucoma detection from segmented images

We use the term logical rule-based frameworks to refer to frameworks that use a set of simple IF-ELSE rules (Figure 4). For such methods to work, the OD and OC are first segmented, then some clinically interpretable imaging features are obtained from the segmented image. Such clinically interpretable imaging features can include variations of the CDR (i.e., vCDR ratio, ACDR and RDR) and measurements from the NRR (i.e., NRR area, area in quadrants, ISNT rule compliance). These features are then used in the framework via IF-ELSE formats for glaucoma classification as presented in (Table 2). In the following text, we reflect on the key aspects of the reviewed papers that apply a logical rule-based AI framework.

[Table 2 here]

**Clinical features used by logical rule-based AI frameworks.** The success of the logical rule-based frameworks is highly dependent on the imaging features used. From the eleven papers identified, eight of the papers used one feature, one paper combined two features, and two papers combined three features for their proposed detection rule (Table 2). The most frequently used feature was the ACDR which was used by six different frameworks. Following this, the vCDR was used by five frameworks, and a variation of the ISNT rule was exploited by two frameworks. The features of vessel ratio index (VRI) and RDR were both used once in combination with other features.

**Logical rule-based AI frameworks using one feature.** Variants of the CDR parameter have proven to be popular due to their clinical value, interpretability and cheap computation from a segmented fundus image. However, some authors have criticised the use of a CDR feature alone, stating that the feature is a limited and incomplete parameter for classifying glaucomatous neuropathy (49,35).

The vCDR was used alone in a detection rule by Dutta et al. (24) with a reported accuracy of 90%. This framework was tested on a small sample of ten images thus, only one image was incorrectly classified (24). The one incorrectly classified image displayed a vCDR of 0.6 which their rule classified as healthy yet the ground truth from ophthalmologists marked the observation as glaucomatous. Although a small study, this example highlights why using the vCDR alone can be problematic. Clinically, it is known that healthy individuals with a large disc can display large vCDR values, and conversely, glaucoma patients with a small disc can have small vCDR values[G]. The authors also recognised this pitfall and propose that future work should consider incorporating other clinically interpretable features.

Two other reviewed papers considered the vCDR alone. Ahmad et al. obtained an accuracy of 92%, sensitivity of 93%, and specificity of 88% (3). Whilst Soorya et al. obtained an accuracy of 97%, sensitivity of 96.5%, and specificity of 98% (50). Both frameworks (50,3) only tested their approach on a dataset acquired from one source which limits the conclusions that can be made about the frameworks' generalisability.

Further work by Dutta et al. (23) proposed the use of the ACDR. The authors stated that the parameter of the ACDR is more appropriate than the vCDR parameter for glaucoma classification. They reasoned that the vCDR parameter assumes that the OC and OD are virtually circular; thus the parameter will not account for any shape irregularities that occur with glaucoma neuropathy.

When using the ACDR alone, the reported accuracies from three papers ranged from 83% (23) to 90% (1) and 96.8% (10) (Table 2). Note, all three papers did not provide the metrics of sensitivity or specificity and used only one dataset. Two other papers (54,58) also used the ACDR parameter alone, Mvoulana et al. (54) yielded an accuracy of 98%, sensitivity of 100% and specificity of 94% and Ong et al. (58) yielded a balanced accuracy of 86% and a sensitivity and specificity of 82% and 89% respectively.

**Logical rule-based frameworks using two or more features.** Rather than using one feature alone, Das et al. (18) proposed combining the vCDR with the ISNT rule for their detection rule. They classified an image as 'healthy' if the vCDR < 0.5 and it satisfies the ISNT rule, otherwise, the image was labelled as glaucomatous. Upon inspection of the framework's misclassifications, they determined that these occurred due to the segmentation step rather than the features used (18). Thus highlighting the importance of accurate segmentation methods in the first step of the framework.

Vijapur and Kunte (80) used the three features of ACDR, rim-to-disc ratio, and vessel ratio index (Table 2). The authors cite that their detection rules were determined after consultations with ophthalmologists to ensure they were clinically relevant and appropriate (80). Their framework introduced the novel idea of segmenting blood vessels and accounting for this within glaucoma classification. However, further external testing is required to evaluate whether the vessel ratio index feature is generalisable to images from other sources.

Three clinically interpretable imaging features: vCDR ratio, ACDR & ISNT rule compliance were used by Issac and Dutta (36), the authors used a logical rule presented in a hierarchical IF-ELSE format (Table 2). This framework resulted in an accuracy of 93%, sensitivity of 94%, and specificity of 96% (36). In frameworks when rules are used in a hierarchical format such as this, it is important to note which features are first in the chain. Whilst it is widely used in practice, the ISNT rule is shown to be less reliable than the vCDR parameter, thus more errors could occur by applying the ISNT rule first (61).

Das et al. proposed the use of vertical cup-to-disc ratio in combination with the ISNT rule (18), the method was tested on four publicly available datasets and one private dataset. This framework resulted in an accuracy of 94%, sensitivity of 92.6%, and specificity of 94.5% (18). Following this, Issac and Dutta used the ACDR parameter with the vCDR parameter and the ISNT rule, yielding an accuracy of 93%, sensitivity of 94%, and specificity of 96% (36). Finally, the paper by Vijapur and Kunte used the ACDR with the RDR parameter and vessel ratio index (80). They obtained a sensitivity of 93% and specificity of 92%; the accuracy of the framework was not provided (80).

## 5.4. Machine learning/statistical modelling – based AI frameworks for glaucoma detection from the segmented image

The machine learning or statistical modelling–based AI frameworks differ from the logical rule-based AI frameworks as they implement a mathematically complex classifier to perform the classification of glaucoma. Alike to the logical rule-based AI frameworks, they can make use of clinically interpretable features extracted from a segmented fundus image. However, different to the logical rule-based AI frameworks they can also create and utilise abstract features and exploit these within machine learning or statistical modelling classifiers. The following section presents the findings of the seventeen papers found in this review that implement a machine learning or statistical modelling-based AI framework.

[Table 3]

### 5.4.1. Machine learning/statistical modelling – based AI classifiers and their reported performance

The machine learning/statistical modelling frameworks differed from one another by the type of classifiers they implemented (Table 3). Support vector machines (SVM's) were the most popular classifier, being used in just over half of the papers (9 out of 17 papers). The clustering methods of M-Mediods and K-nearest neighbours (K-NN) were used by one paper each and the ensemble classifier of Random Forest (RF) was also proposed once. Additionally, one paper used Linear Mixed Effects (LME) modelling. The remaining four frameworks proposed a variant of a neural network (NN) for classification. Note that, Table 3 only presents the optimal classifier used in the frameworks. That is, many papers propose a range of classification models/algorithms and present the classifier which worked optimally.

**Support Vector Machine Classifiers.** A total of nine papers used support vector machine classifiers, and two different kernel functions were selected within these. The radial basis function (RBF) kernel was used by five frameworks and the linear kernel was used by three frameworks. From the papers using the RBF kernel, Issac et al. (37) obtained an accuracy of 94%, sensitivity of 100%, and specificity of 90%. Krishnan et al. only provided the F1 score as a metric, which was 91% (44). The framework proposed by Agarwal et al. obtained an accuracy of 90%, sensitivity of 100% and specificity of 80% (2); whilst the framework by Khalil et al. combined two support vector machines with RBF kernels and achieved an accuracy of 92.9%, sensitivity of 87.5% and specificity 90.84% (42). Khalil et al.

found significant improvement in classification capabilities was achieved by combining the outputs of the support vector machine classifiers and considering a range of structural and textural features. A more recent method by Kang et al. resulted in an accuracy of 85.06%, sensitivity of 81.95% and specificity of 88.28% (39).

From the three papers that used support vector machine classifiers with linear kernels, Narasimhan and Vijayarekha only provided the metric of accuracy which was 95% (56). Mukherjee et al. obtained an accuracy of 87%, sensitivity of 86.4% and specificity of 90% (52). More recently, Pathan et al. achieved an accuracy of 96.66%, sensitivity of 100% and specificity of 95% with the publicly available DRISHTI database but on external testing with a private database, this reduced to an accuracy of 90%, sensitivity of 93.47% and specificity of 91.2% (59). Deepika and Maheswari did not specify the kernel used, this framework yielded an accuracy of 91.67%, sensitivity of 90% and specificity of 93.3% (21).

**Clustering classifiers.** Clustering methods were used by two frameworks. The k-nearest neighbours' algorithm (K-NN) was proposed by Lotankar et al., achieving an accuracy of 99.2%, sensitivity of 86.7% and specificity of 84% (48). The framework by Akram et al. used a clustering method of M-Medoids (5). They proposed that there is variation in the number and distribution of the samples within the two classes (healthy & glaucomatous) and via employing multivariate m-modelling and classification, they could handle multimodal distribution of samples within the two classification groups (5). This method was tested on five datasets; the accuracy across the datasets ranged from 86.7 – 94.4 %, sensitivity from 75 – 93.3% and specificity from 87.1 – 97.1% (5).

**Random Forest classifier.** A Random Forest classifier was proposed by Zahoor and Fraz (81). This method resulted in an accuracy of 95.3%, sensitivity of 96.31% and specificity of 95.33%. However, the authors state the use of the publicly available High-Resolution Fundus Image (HRF) database but removed nine of the total 36 fundus images without explanation.

**Linear mixed-effects statistical modelling.** A linear mixed-effects (LME) modelling approach was used by one paper (49). This framework by MacCormick et al. yielded an AUROC of 99.7%, sensitivity of 100% and specificity of 98.3% on internal testing. The proposed framework then employed external validation using the publicly available RIM-ONE V3 dataset, the AUROC obtained was 91% (49). A disadvantage of such an approach is in needing the segmented image of healthy eyes to follow a statistical model with a plausible number of parameters. This is not always possible, however, in the case of glaucoma, this was a suitable approach. The authors determined that the contours of OD and OC appeared to be two-centred ellipses in healthy eyes and additionally, they included a technique to account for each eye displaying different disc sizes – all of which was captured in the statistical model. Using this information, the classification of glaucoma was then based on a deviation of the contours from the model of healthy eyes.

**Neural network classifiers.** A multi-layer perceptron was proposed by Perdomo et al., with the final stage being composed of two fully connected layers with 64 hidden and 3 output units (60). The batch size, the number of epochs and optimal features were determined via a grid search. The binary classification achieved an accuracy of 89.3%, sensitivity of 89.5%, specificity of 88.9% and AUC 82%. Using multi-class classification (healthy, suspected glaucoma and glaucoma), they provided the metrics of precision and recall which were 0.76 and 0.72 respectively.

Raja and Ramanan proposed the use of Damped Least-Squares Recurrent Deep Neural Learning Classification (DLRNL) (63). The classification was performed on the output layer using soft sign activation functions resulting in an accuracy of 89% however, no other performance metrics were specified. The paper by Karkuzhali and Manimegalai (40) considered a range of networks, the best performance was found when using the Feed Forward Back Propagation Neural Network (FFBPNN) and the Distributed Time Delay Neural Network (DTDNN); each of these provided an accuracy of 100% and sensitivity and specificity of 100%. Note that they tested on a small subsection of the publicly available DRISHTI dataset consisting of just 26 images. Kausu et al. used a multi-layer perceptron and obtained an accuracy of 97.67%, sensitivity of 98% and specificity of 97.1% (41). Note they did not provide any detail of the multi-layer perceptron (i.e., number of neurons in each layer, hyperparameter tuning etc).

### 5.4.2. The machine learning/statistical modelling AI frameworks utilise clinically interpretable image features as well as abstract image features

The machine learning/statistical modelling-based AI frameworks reviewed used clinically interpretable and abstract image features (Table 3). Across all the papers reviewed, each framework used some variant of the CDR parameter, highlighting the significance of the parameter in glaucoma classification. Ten of the papers used clinically interpretable imaging features (i.e., vCDR ratio, NRR area etc), 7 papers proposed the use of novel spatial/spectral/texture/colour features.

The use of spatial features by Akram et al. (5) was motivated by the fact that the area of the OC changes from the normal to the glaucomatous eye. In keeping with MacCormick et al. (49), they state the use of the vCDR parameter alone was limited due to glaucoma manifesting at any direction in the ONH. Whereas Akram et al. combined the RDR parameter with spatial and spectral features (5), MacCormick et al. proposed using a profile CDR (pCDR) which quantifies the optic nerve rim consistency around the whole disc at 15-degree intervals (49). Moreover, due to the use of linear mixed-effects modelling by MacCormick et al., random effects were incorporated to indirectly take account of the size of the OD.

The framework by Kausu et al. (41) exploited clinically interpretable imaging features and abstract features in combination. Wavelet features were considered as the authors argued that texture features alone are not enough, as they do not consider frequency information. Yet, by exploiting the wavelet transform, frequency and spatial information would be considered. Kausu et al. (41) used the minimum redundance maximum relevance (mRMR) feature selection technique to determine the optimal features from the collection of clinically interpretable and wavelet features. However, in the end, the best performance was obtained when only using two features: vCDR and the second-order texture feature - energy. Whilst the vCDR parameter is clinically interpretable, the second-order feature of energy is an abstract feature.

Correlation-based feature selection was applied by Pathan et al.(59) ; they began with 54 colour features, 12 texture features and 2 clinically interpretable features. Following feature selection, 10 features (2 clinical, 4 colour & 4 texture) were deemed relevant and applied in the final framework.

Mukherjee et al. (52) proposed a framework using eight features (Table 3). They compared this framework with the same methodology but using only the CDR feature yet, they found this resulted in significantly lower performance metrics. Thus, indicating the relevance of the other parameters used. However, this is to be further examined to test the generalisability of the other features for glaucoma classification with external datasets (52). Similarly, Khalil et al. (42) found improved performance by combining structural and textural features for classification – this achieved high performance metrics (Table 3).

## 5.5. Approaches to segmentation

Intuitively, the success of a multi-step framework depends on the type and success of the automated ONH segmentation used in the first step of the framework. Although the focus of this review is not to assess the automated segmentation methods, in this section we give a brief overview of the approaches to segmentation used. Briefly, some automated segmentation methods focus on colour intensity and texture-based thresholding. Some advanced methods employ fully convolutional neural networks. The point is that there are many different approaches to segmentation with differing degrees of success. In the segmentation of the ONH, it is well-known that the OC is much more challenging to segment than the OD due to the low contrast between the

OC and the neighbouring disc region (23). As such, there are very few papers focused on developing OC segmentation methods.

## 5.6. Glaucoma disease groups

From the 28 papers identified in this review, 24 (85%) papers performed binary classification (healthy or glaucoma), and 4 papers performed multi-class classification (healthy or suspected glaucoma or glaucoma). Across the four papers performing multi-class classification, their method for incorporating the suspected class differed.

The framework by Khalil et al. (42) used a combination of clinically interpretable and abstract features in two support vector machine classifiers (one support vector machine using structural features and one support vector machine using textural features) for glaucoma classification. They proposed that if the outputs of the two support vector machine classifiers did not agree (i.e., one support vector machine provides the outcome healthy and the other glaucomatous) they would classify this image as 'suspect glaucoma'.

Perdomo et al. (60) proposed a multi-layered perceptron with 3 output units using 19 morphometric features. They used the publicly available RIM-ONE V3 dataset which comprises 35 suspected glaucoma fundus images for training/testing their framework to handle the suspected class. Although they showed high performance metrics on binary classification, the performance on multi-class classification was inferior; the metrics of precision and recall were 0.76 and 0.72 respectively (60). Thus, their framework was not optimal when considering the suspected glaucoma class.

More recently, frameworks by Soorya et al. (50) and Issac and Dutta (36) applied logical rule-based AI frameworks with thresholds for glaucomatous and healthy; if the features obtained from the segmented fundus image did not meet the criteria for the glaucoma or healthy group, this was classified as suspected glaucoma (Table 2).

## 5.7. Approaches to validation of methods and the reporting of performance metrics

**The approach to validation in logical rule-based AI frameworks.** In the papers that used a logical rule-based AI framework, the approach to validation differed as they have no training component within their frameworks. The only means of validation per se (using a logical rule-based framework) is to acquire datasets from a range of sources to evaluate if their proposed rules are generalisable/appropriate. From the 11 logical rule-based frameworks (Table 2), nine papers (81%) used one dataset, one paper used two datasets, and one paper used 5 datasets. As such, the majority of papers using logical rule-based frameworks did not consider validation of their proposed frameworks.

Considering the performance metrics presented, six papers (54%) presented the performance metrics of accuracy, sensitivity & specificity whilst the remaining 5 papers did not. Four of the papers only provided their accuracy result, and one paper did not provide accuracy, only sensitivity & specificity.

**The approach to validation in machine learning/statistical modelling AI frameworks.** The machine learning/statistical papers differed in their approach to framework validation. The approach of 10-fold cross-validation was used by five papers, 5-fold cross-validation was used once, and leave-one-out cross-validation was also used once. The remaining papers used a data split for validation. A 70:30 split was used three times whilst a 50:50 split was used twice. External validation was only used by one paper. Four papers used more than one database within their frameworks. For this, they either trained and tested their model individually on the different databases or they combined the databases and then trained and tested on the data (Table 3).

In addition to conducting internal/external validation, some of the reviewed papers compare their AI method with previously published methods. Nine of the 17 machine learning papers compared their proposed methodology with previously published methods. Whilst 13 papers compared their methods with at least one other method proposed by themselves.

Considering the performance metrics reported, 14 (82%) papers disclosed metrics of accuracy, sensitivity and specificity. Only two papers presented metrics for AUC and one paper presented no metrics other than the F1 score. Additionally, two papers only presented the accuracy metric results.

## 5.8. Databases used for development and testing of the AI frameworks reviewed

Within the frameworks highlighted in this review, a range of publicly available and private databases were used, an overview is provided in Table 4.

[Table 4 here]

### 5.8.1. Publicly available datasets

**DRISHTI dataset.** From the 28 papers identified, the DRISHTI database (70) was one of the most popular databases being used seven times. The database comprises 101 fundus images (31 healthy and 70 glaucoma) acquired at Aravind Eye Hospital, Madurai, India. This dataset is of a single population as collected images are from subjects who are Indians. The images were taken with the eyes dilated using the following data collection protocol: centred on the OD with a Field-of-View of 30-degrees and dimension 2896 × 1944 pixels. Low-quality images (poor contrast, positioning of OD region, etc) were not used. The ground truth for the region boundaries, segmentation soft maps and CDRs by 4 different ophthalmologist experts (with varying clinical experience) is provided. The database is split into 50:51 training: testing. Note that, to access the ground truth for the test data, a researcher must register with the data owners (69).

**High-Resolution Fundus (HRF) dataset.** The HRF dataset (11) was used by seven of the reviewed papers. In comparison to the other databases available, this database is small, comprising 45 fundus images in total. The images were collected at the same clinic in the Czech Republic (57). Of the 45 images, 15 are glaucomatous, 15 healthy and 15 are labelled as diabetic retinopathy. The database is publicly available and in an easily downloadable format online. All fundus images were acquired with a mydriatic fundus camera CANON CF–60 UVi equipped with a CANON EOS–20D digital camera with a 60-degree field of view (FOV). The image size is 3504 × 2336 pixels (57). The database curators do not state how many ophthalmologists were used for the ground truth. As well as the fundus images, researchers can access the Field Of View (FOV) masks, vessel segmentation, and OD gold standards provided by three experts (57). Whether the images were obtained in a dilated state or not is not disclosed.

**Messidor dataset.** The Messidor database (20) was used in two reviewed papers. It contains a total of 1200 images of different diseases, but only 100 images are annotated for glaucoma. From the 100 fundus images, 28 are glaucomatous and 72 are healthy. The images were acquired by 3 ophthalmologic departments in France using a colour video 3CCD camera mounted on a Topcon TRC NW6 non-mydriatic retinography with a 45-degree field of view. To access the dataset, the researcher is required to submit a form that is evaluated by the data owners and they decide upon the validity of the request and provide permission (20).

**ORIGA dataset.** The ORIGA database (82) was used in one reviewed paper. The ORIGA database consists of 650 fundus images in total, 168 glaucomatous images and 482 randomly selected non-glaucoma images. The authors state that there are 336 images from the left eye and 314 from the right. The ORIGA database was formed using retinal image data collected from the Singapore Malay Eye Study (SiMES) (26) conducted by the Singapore Eye Research Institute. Each image is tagged with grading information (CDR, ISNT rule compliance, RNFL defects, Notches and PPA) and the manually segmented result of the OD and OC (82). Although it is stated that it is publicly available, it is not easily accessible from searching online. Moreover, no details are provided regarding the imaging device used (43).

**RIM-ONE dataset.** Four reviewed papers utilised the RIM-ONE databases (8,27,28). There are three different versions of the RIM-ONE databases: V1 and V2 which were used once and V3 - which was used twice. RIM-ONE V1 (27) was published in 2011; the dataset comprises 169 fundus images from different subjects. There are 5 groups: Normal, Early Glaucoma, Moderate Glaucoma, Deep Glaucoma and OHT (Ocular Hypertension) which have 118, 12, 14, 14 and 11 images respectively. The RIM-ONE V1 database consists of 5 manual reference segmentations per image. This enables the creation of reliable gold standards, thus decreasing the variability among expert segmentations and the development of highly accurate segmentation algorithms (27). The fundus images were acquired from three different hospitals located in different Spanish regions (Hospital Universitario de Canarias, Hospital Clínico San Carlos and Hospital Universitario Miguel Servet). The authors of RIM-ONE state that compiling images from different medical sources guarantees the acquisition of a representative and heterogeneous image set (27). The images were captured using a Nidek AFC-210 non-mydriatic fundus camera with a 21.1-megapixel Canon EOS 5D Mark II body, with a vertical and horizontal field of view of 45°.

The RIM-ONE V2 dataset (28) comprises 455 fundus images (200 glaucomatous and 255 healthy), the ground truth for the images were provided by one expert ophthalmologist. The most recent version of the database is the RIM-ONE V3 which includes 159 fundus images with 85 healthy, 39 glaucoma and 35 suspected glaucoma. The images were taken by a non-mydriatic Kowa WX 3D stereo fundus camera (2144 × 1424 pixels) and 34-degree POV. The images were acquired at the Hospital Universitario de Canarias and the ground truths provided by two experts (8).

**GlaucomaDB dataset.** The GlaucomaDB (42) database was used twice by frameworks in this review. The database is a subset of 120 fundus images from a larger database of 462 images gathered in a local hospital. The region/country of the local hospital was not disclosed. The images were captured using a TopCon TRC 50EX camera with a resolution of 1504 × 1000. The 120 images consist of 85 healthy and 35 glaucomatous, with the ground truths provided by two ophthalmologists (42). To access the database for research purposes, permission from the authors has to be requested.

**HEI MED dataset.** The HEI Med Dataset (Hamilton Eye Institute Macular Edema Dataset) (31) is a collection of 169 fundus images, however, only 50 images are annotated for glaucoma detection. The HEI MED database was used by one framework. Of the 50 images, 30 are healthy and 19 are glaucomatous. The fundus images were collected at the Hamilton Eye Institute, United States of America, via a Visucam PRO fundus Camera (Zeiss) (43) and annotated by one ophthalmologist. The data is available on GitHub for public use.

**DRIONS dataset.** The DRIONS database was used twice by papers in this review. The database comprises 110 fundus images (95 healthy and 15 glaucomatous). The images were collected at the Ophthalmology Service at Miguel Servet Hospital, Saragossa, Spain. Images were removed if any form of cataracts were present. All images were obtained from subjects of Caucasian ethnicity. The images were acquired with a colour analogical fundus camera, approximately centred on the ONH and they were stored in slide format. To have the images in digital format, they were digitised using a HP-PhotoSmart-S20 high-resolution scanner, RGB format, resolution 600 x 400 and 8 bits/pixel (13). The dataset is easily accessible and is available to download online.

**DIARETDB dataset.** The DIARETDB [H] database consists of 89 colour fundus images and was primarily developed for aiding diabetic retinopathy research. However, the database has been made publicly available and it has been assessed for glaucoma. The 89 fundus images are split into 81 healthy and 8 glaucomatous (5), four medical experts were collected for the ground truth annotations. All images were captured using the same 50-degree field-of-view digital fundus camera with varying imaging settings at Kuopio University Hospital, Finland. The database is easily accessible for download online.

**DRIVE dataset.** The DRIVE database was used by one paper in this review. The database comprises 40 fundus images (34 healthy and 6 glaucomatous) annotated by two ophthalmologists (42). The images were acquired using a Canon CR5 non-mydriatic 3CCD camera with a 45-degree field of view (FOV). Each image was captured using 8 bits per colour plane at 768 by 584 pixels. Although stated that the database is publicly available, it is not easily accessible online.

### 5.8.2. Private datasets

Private databases were popular in the development (training and testing) of the frameworks reviewed, a total of 15 private databases were used. However, due to not being publicly available it was difficult to access detailed information on the databases if not provided directly from authors. Many papers did not provide basic information other than the dataset size. Without all information regarding the datasets (i.e., patient cohort, imaging device, etc), it is difficult to draw conclusions regarding the robustness and generalisability of the proposed frameworks (as this is dependent upon the dataset used).

## 6. Discussion and conclusions

We present the first review, to our knowledge, of AI frameworks for glaucoma detection that utilise fundus images and produce ONH segmentation as the first step. By segmentation, we refer to images that have been automatically partitioned into three areas: optic cup, optic disc and neuroretinal rim. We have specified such inclusion criteria due to the benefits of using segmented fundus images within glaucoma detection strategies.

We focused on the modality of fundus imaging as it is the simplest modality of ONH assessment. The quality of fundus images may be sufficient for evaluating ocular health for the presence of glaucomatous neuropathy and due to its relatively low cost, fundus cameras are readily available in a range of settings. As such, there is the potential to exploit fundus images via AI to develop automatic glaucoma screening provisions, even for economically weak areas in the world. Additionally, the segmented images have inherent interpretability and explainability – one can explain how the AI works and why the AI thinks that the eye is glaucomatous or not.

Within this review, we aimed to identify where we are with exploiting the segmented fundus images for automatic glaucoma detection and to discuss how we can move forward to progress in the field. We found 28 papers published between 2011 and March 2021. Between 2011 and 2019 there has been growing active research in this area, then in 2020, the number of published papers was lower.

### 6.1 Three key findings of this review

#### 6.1.1 Glaucoma detection via two-step AI frameworks using fundus images present encouraging results

We found that the two-step AI frameworks have presented promising results in their first step when identifying the contours of OC and OD (i.e., segmentation of the ONH). We then identified two approaches to using features derived from the segmented fundus images: logical rule-based frameworks and machine learning/statistical modelling frameworks.

This review highlights that the glaucoma detection performance of the logical rule-based AI frameworks is limited due to the nature of using simplistic rules (even more so when the simplistic rules have been derived from small homogenous datasets). We found that ten papers split one dataset for training and testing and reported accuracy ranged from 83 to 97%. Since this accuracy was determined via internal validation and on small datasets it must be interpreted with caution.

One paper (Vijapur and Kunte (80)) did perform external testing (i.e., they used more than one data source). They reported two combinations of sensitivity vs specificity: 93 vs. 92%, and 87 vs. 87%. Across all 11 papers, we found that there was no consensus on thresholds applied within the rules for glaucoma classification. That is, although many papers highlighted that their rule was based upon clinical relevance (as they were using a clinical parameter within their rule i.e., vCDR), the threshold used for the clinical parameter changed from one paper to another. Consequently, this highlights that a given threshold may only be appropriate for the dataset at hand. Moreover, as the majority of the logical rule-based AI frameworks did not implement any external testing, we are limited in understanding how the framework would work in screening strategies with data collected from different sources.

Regarding the machine learning/statistical modelling-based AI frameworks reviewed, we found that the reported accuracy was between 85.1 and 100%, predominantly reported via internal validation. The reported performance of some of the frameworks was comparable to that of the one-step approaches using DL. One of the current DL approaches is by Li et al.(46) who reported an accuracy of 0.986. However more direct comparisons are required on the same datasets to give a fair comparison of the approaches.

### 6.1.2 There is active research into developing two-step AI frameworks for glaucoma, where the first step is automatic detection of OC and OD contours

We conducted this review by focusing on two-step AI frameworks that produce ONH segmentation as a first step. One key reason for this is the interpretability and explainability benefits that can be found when using segmented images within AI frameworks. It is known that the segmented contours of OC and OD can explain to the clinician why the AI has classified a given fundus image as glaucomatous or not. The two-step solution helps visually explain intermediate steps between the raw image and diagnosis (19). This can significantly aid in the development of trust within the AI and consequently the adoption of such methods within the practice of glaucoma detection. Moreover, such explainable AI methods can act as a support decision system such that the AI, clinician and patient can work together to decide upon treatment options and next steps.

A principal advantage of the reviewed two-step AI frameworks is that they require smaller datasets than one-step frameworks that utilise DL. For example, the two-step framework by MacCormick et al. (49) achieved an of AUROC 99.6% and 91.0%, in internal and external validation respectively, while using approximately 300 images for training. Whilst a one-step DL framework proposed by Li et al. (46) achieved comparable accuracy but used 30,000 images for training. Two-step approaches can require 100 times fewer data as they are 'simpler' methods with few inputs in comparison to DL methods. The use of the whole fundus image in DL methods means that the methods have a large amount of data to handle, much of which is redundant – with the most important information appearing to lie in the boundaries of the ONH. In areas outside of ophthalmology, it has also been observed that neural networks can be made more data-efficient if they utilize feature contours, for further detail see the work by George et al (30).

### 6.1.3 Colour fundus images are actively studied for their potential use within AI-enabled glaucoma detection

This review solely focuses on glaucoma detection frameworks using fundus imaging technology. This choice was guided by the fact that the detection of glaucoma in clinical practice is highly influenced by optic nerve head assessment via fundus imaging and the use of colour fundus is part of the guidelines for glaucoma diagnosis. Moreover, colour fundus images are advantageous due to their lower cost in comparison with other imaging modalities and the technology is continuously developing such that they can consistently provide high-quality images capable of distinguishing glaucomatous neuropathy.

It should be highlighted that there is a distinction between large fundus cameras, costing 10s of thousands of pounds, and the recently developed smaller mobile phone cameras that enable fundus imaging of the ONH at a considerably lower cost. While other imaging modalities such as OCT can provide additional information and are becoming more widely available, it is currently hard to see if lower-cost mobile OCT is possible and hence whether it will be available to less developed countries and remote areas. Yet portable fundus cameras are becoming increasingly accessible and viable, even within economically less fortunate countries.

## 6.2 Three key unresolved issues of current knowledge and potential areas for future studies

### 6.2.1 There is a need to work on AI frameworks that utilise colour fundus images and that provide contours of OC and OD in their first step

A direct comparison of all approaches for AI-enabled glaucoma detection methods is required. One-step AI approaches (end-to-end approaches, based on DL) need to be compared to two-step approaches reviewed here, on the same datasets. This will ensure a direct comparison can be made and one can consequently identify the benefits and drawbacks of both approaches. For now, we can identify that the advantage of the one-step AI is that it does not require such a large effort in terms of segmenting the fundus images and deriving clinical features from the segmentations; this is due to the nature and complexity of DL. However, a major disadvantage is in the lost interpretability (due to the black-box nature of DL) and in needing many annotated images to be trained. Such algorithms need to be studied together with two-step algorithms, to understand better which are more suited for glaucoma detection.

Furthermore, more research needs to be done on the comparison of imaging modalities. Specifically, investigations need to be made between OCT and fundus imaging to comprehensively compare both modalities such that we can determine which is most suitable for AI-enabled glaucoma detection frameworks.

Moreover, further research should be conducted regarding the development of AI-enabled glaucoma detection frameworks. For example, an area to be studied is the quantification of uncertainty of the outputs provided by the AI framework$^D$. Also, the inclusion of other data sources needs to be investigated, e.g. patients' de-identified personal data, genetics data, visual fields data. This will simulate the clinical workflow as well as potentially improve the performance of the AI frameworks and help to explain AI outputs.

### 6.2.2 There is a need to keep building and sharing suitable datasets

There exist several large landmark clinical study datasets which were not used in the publications reviewed - despite being a very rich resource of clinical images for glaucoma diagnosis. This includes the United Kingdom Glaucoma Treatment Study (UKGTS) (29), the Ocular Hypertension Treatment Study (OHTS) (32), and the Northern Ireland Cohort Longitudinal Study of Ageing (NICOLA) (51). There are several possible reasons for the exclusion of such datasets. These datasets lack pixel-level image annotation of the OC and OD, which is required to train and validate segmentation models. Acquiring such annotations is a time-consuming task requiring expert knowledge. Furthermore, access to these datasets requires an application, payment and submission of a suitable protocol, which can act as barriers.

Moving on, our review has highlighted the increasing need for datasets to include the whole spectrum of glaucoma severities (not just glaucoma and normal, but also for glaucoma suspects). This is crucial to the development of AI frameworks that are useful in clinical practice as 'suspect glaucoma' is a case regularly observed by clinicians. Additionally, it is very important to collect and develop resource-rich longitudinal datasets such that disease onset and progression can be examined and incorporated into AI frameworks.

We also highlight that it is essential that sufficient details are provided alongside datasets. This includes the number of patients, the number of images acquired from each patient, whether both eyes are used (i.e., an image per eye) etc. All of this information is important and relevant for researchers developing AI frameworks as such methods can be based upon hard assumptions. If these assumptions are not upheld due to the lack of information provided with the datasets, this can cause significant issues. Additionally, other information that should be recorded includes the type of camera used, number of ophthalmologists for annotation of images/providing of ground truth, source of data, inclusion/exclusion criteria for data collection etc was also limited. In particular, when data has been acquired from a private source, there has been a scare amount of information provided. A detailed description of the dataset used is critical for the assessment of the quality, reliability, suitability to produce the desired output, potential generalisability of any findings, and especially reproducibility of the methods (74).

Another important point to highlight is that further effort is required to ensure datasets are provided with suitable gold standards (aka ground truth). High-quality gold standards are crucial for AI development. A means of achieving this is acquiring annotations from multiple human graders. The reasoning for this is that ONH annotation (via fundus images) can suffer from large amounts of inter-observer variability – it is a subjective task (38,14). Using only one grader introduces bias into the ground truths upon which the AI is developed. A useful measure of the reliability of ground truth labels is inter-observer agreement between the labellers. By detailing the inter-observer agreement, readers can make a judgement on the likelihood that the ground truth label is correct. This review has highlighted that only 3 of the 12 publicly available databases have more than two annotators. Whilst it should be standard to have more than 2 annotators, it should be recognised that obtaining manual annotation of images is not an easy task as it is time-consuming, expensive and requires expertise.

Moreover, further work is required to improve the diversity of datasets. The use of the term diversity here refers to having fundus images captured from various devices, involving different patient ethnicities, and images taken in different lighting, contrast and noise (8). The frameworks reviewed here are developed on datasets predominately acquired from one source and as such lack this diversity. A potential limitation of this is whether the quoted sensitivities and specificities will be generalisable to real-world patient cohorts where a range of factors can negatively impact the quality of the fundus images (53). Moreover, selection bias can be present if the dataset has been collected from homogeneous sources (i.e., using one ethnicity and/or specific hardware/imaging settings). Methods developed on such datasets are prone to generalisation problems as one population data might have different characteristics that introduce bias in the proposed framework (67).

### 6.2.3 There is a need to keep developing guidance for high-quality reporting of AI frameworks and promote following the guidance

This review highlights that several publications lacked high-quality reporting – both in terms of datasets used and their glaucoma classification methodology. Some of the reviewed papers lacked the technical details regarding their classifier whilst others only provided a brief explanation of the methods selected. Often lacking sufficient detail was the model structure (i.e., hyperparameters used and their tuning mechanism). The limitation of not providing sufficient details of methods (i.e., technical AI details) is that this renders the paper unreproducible, a key criticism in the AI field. There is a need to support the work of the initiative EQUATOR [F] which is a collaboration between experts in statistics, machine learning and computing – but it also involves specialised clinicians and policymakers. This initiative develops and provides detailed guidance for reporting, with a specific focus on guidance for medical studies involving AI.

# 7. Method of Literature Search

We used four databases to search for relevant literature: PubMed, Scopus, Web of Science and Medline (OVID). The search covered all years up until 2021. The search strategies are detailed in supplementary file 1.

## Search terms

### Database: Scopus

( TITLE-ABS-KEY ( glaucoma )  AND  TITLE-ABS-KEY ( fundus  OR  retinal )  AND  TITLE-ABS-KEY ( classif*  OR  discrim*  OR  diagnos* )  AND  TITLE-ABS-KEY ( photograph*  OR  imag* )  AND  TITLE-ABS-KEY ( "auto* detect*"  OR  "detect"  OR  "predict*" )  AND  TITLE-ABS-KEY ( segment* ) )

### Database: PubMed

(((((glaucoma[Text Word]) AND (fundus[Text Word] OR retinal[Text Word])) AND (classif*[Text Word] OR discrim*[Text Word] OR diagnos*[Text Word])) AND (photograph*[Text Word] OR imag*[Text Word])) AND ("auto* detect*"[Text Word] OR detect*[Text Word] OR predict*[Text Word])) AND (segment*[Text Word]

### Database: Web of Science

TS = (glaucoma AND (fundus OR retinal) AND (classif* OR discrim* OR diagnos*) AND (photograph* OR imag*) AND ("auto* detect*" OR detect* OR predict*) AND (segment*))

### Database: MEDLINE

(glaucoma and (fundus or retinal) and (classif* or discrim* or diagnos*) and (photograph* or imag*) and (auto* detect* or detect* or predict*) and segment*).tw.

## Eligibility criteria

### We included papers if:

1. The paper uses segmented fundus images of the Optic Nerve Head (ONH).
2. The paper proposes a methodology/framework for the classification of glaucoma.
3. Full text is available online.
4. Full text is available in English.

### We excluded papers if:

1. Interested purely in segmentation of fundus images (provide no classification of glaucoma following segmentation).
2. Focused purely on classification via methods that require no segmentation of the fundus image (i.e., one step AI frameworks).

# 8. Disclosure

This work was supported by a PhD studentship funding from the British Council for Prevention of Blindness to Dr. Gabriela Czanner, Dr. Silvester Czanner, Professor Colin E. Willoughby and Dr. Srinivasan Kavitha.

## 9. Permission Requirements

Not applicable.

## 11. Other Cited Material